\documentclass{article}

\PassOptionsToPackage{numbers, compress}{natbib}

\usepackage[preprint]{neurips_2021}

\usepackage[utf8]{inputenc} 
\usepackage[T1]{fontenc}    
\usepackage{natbib}
\usepackage{url}            
\usepackage{booktabs}       
\usepackage{amsmath,amssymb,amstext,bbm,bm} 
\usepackage{mathtools}
\usepackage{nicefrac}       
\usepackage{microtype}      
\usepackage{xcolor}         
\usepackage{graphicx}
\usepackage{subcaption}

\newcommand{\R}{\mathbb{R}}
\DeclarePairedDelimiter\norm{\lVert}{\rVert}

\title{SymbolicGPT: A Generative Transformer Model for Symbolic Regression}

\author{%
  Mojtaba Valipour \thanks{David R. Cheriton School of Computer Science} \\
  University of Waterloo\\
  \texttt{mojtaba.valipour@uwaterloo.ca} \\
  \AND
   Bowen You \footnotemark[2] \\
  University of Waterloo \\
  \texttt{byyou@uwaterloo.ca} \\
   \And
   Maysum Panju \thanks{Department of Statistics and Actuarial Science} \\
  University of Waterloo \\
  \texttt{mhpanju@uwaterloo.ca} \\
   \And
   Ali Ghodsi \footnotemark[1] \hspace{0.1pt} \footnotemark[2] \\
  University of Waterloo \\
  \texttt{ali.ghodsi@uwaterloo.ca} \\
}

\begin{document}

\maketitle

\begin{abstract}
  Symbolic regression is the task of identifying a mathematical expression that best fits a provided dataset of input and output values. Due to the richness of the space of mathematical expressions, symbolic regression is generally a challenging problem. While conventional approaches based on genetic evolution algorithms have been used for decades, deep learning-based methods are relatively new and an active research area. In this work, we present SymbolicGPT, a novel transformer-based language model for symbolic regression\footnote{Code and results available at https://git.uwaterloo.ca/data-analytics-lab/symbolicgpt2}. This model exploits the advantages of probabilistic language models like GPT, including strength in performance and flexibility. Through comprehensive experiments, we show that our model performs strongly compared to competing models with respect to the accuracy, running time, and data efficiency.
\end{abstract}

\section{Introduction}

Deep learning and neural networks have earned an esteemed reputation for being capable tools for solving a wide variety of problems over countless application domains. Notably, deep language models have made an enormous impact in the field of linguistics and natural language processing. With the advances in technology like Generative Pre-trained Transformers, or GPT \cite{gpt1}, the scope of problems now accessible to automated methods continues to grow. It is particularly interesting when language models are used for tasks that, at first glance, do not seem to have any relationship with language at all.

Symbolic regression, the problem of finding a mathematical equation to fit a set of data, is one such task. The objective of symbolic regression is to obtain a closed-form symbolic mathematical expression to describe the relationship between specified predictor and response variables, where the mathematical expression is allowed to be flexible without being restricted to a particular structure or family. More precisely, the goal in symbolic regression is to recover a mathematical function $f$ in terms of the input variables  $\bm{x}= [x_1 \ldots  x_d]^\top$, given a set of datapoint vectors of the form $D = \{(\bm{x}_i, y_i)\}_{i=1}^n$, such that 
$f(\bm{x}_i) = y_i$
for all $i$. Here,  $x_1, \ldots, x_d, y_i$ are scalars and $\bm{x}_i \in \R^d$.

By not imposing any structural constraints on the shape of the desired equation, symbolic regression is a much more difficult task compared to other kinds of regression, such as linear regression or multinomial regression, as the search space of candidate expressions is so much larger. 

The most common approach for symbolic regression is based on genetic programming, where numerous candidate parse trees are generated, evaluated, combined, and mutated in an evolutionary way until a tree is produced that models an expression that fits the dataset up to a required accuracy level. In essence, it is a search strategy over the vast space of mathematical expressions, seeking the formula that would optimize an objective function. 

In this typical framework, which applies not only to genetic methods but also many deep-learning-based approaches to symbolic regression, the goal is to identify a mathematical expression that fits most optimally given a single input dataset. This dataset is the basis over which all the training occurs. Consequently, when presented with any new dataset (as a fresh instance of the task of symbolic regression), the entire training procedure must begin again from scratch.

In this work, we explore an alternative approach to symbolic regression by considering it as a task in language modelling. Symbolic mathematics behaves as a language in its own right, with well-formed mathematical expressions treated as valid ``sentences'' in this language. It is natural, therefore, to consider using deep language models to address tasks involving symbolic mathematics.

We can frame the regression problem as an exercise in captioning. Each instance takes input in the form of a cloud of points in $\mathbb{R}^{d+1}$, with each point consisting of $d$ components corresponding to $\bm{x}$ and a single component for the associated $y$ value. The instance returns a statement in the language of symbolic mathematics to describe the point set. By training a model to correctly ``caption'' datasets with the equations underlying them, we will have a system for performing symbolic regression quickly and accurately. 

Based on this idea, we present SymbolicGPT, a method that makes use of deep language models for symbolic regression. SymbolicGPT employs a framework that represents a major shift from the way symbolic regression is conventionally performed. We move the task of symbolic regression from being a strictly quantitative problem into a language one. Effectively, we propose a system that not only learns the language of symbolic mathematics, but also the underlying relationship between point clouds and mathematical expressions that define them.

As part of SymbolicGPT, we make use of a T-net model \cite{qi2017pointnet} to represent the input point cloud in an order-invariant way. This allows us to obtain vector embeddings of the entire input dataset for symbolic regression instances without depending on the number of points in the dataset or the order in which they are given.

A major advantage of SymbolicGPT is that we are no longer training a model to learn an equation for an individual dataset in each instance of symbolic regression. Instead, we train a single language model once, and use that trained model to rapidly solve instances of symbolic regression as individual captioning tasks. We will show that SymbolicGPT not only presents a running time speed boost of an order of magnitude or more, but also provides competent performance in accurately reconstructing mathematical equations to fit numerical datasets, presenting a new frontier for language models and a novel direction for approaching symbolic regression.

\section{Related Work}
\label{Related Work}

Traditionally, the problem of symbolic regression has been tackled with methods based on genetic algorithms \cite{mckay1995using,augusto2000symbolic,schmidt2009distilling,murari2014symbolic,wang2019symbolic}. In this framework, the task is seen as a search space optimization problem where symbolic expressions are candidates and the expression with the greatest fitness, or fitting accuracy on the training data, is obtained through a process of mutation and evolution. Although this approach has shown success in practice, it is computationally expensive, highly randomized, requires instance-based training, and struggles with learning equations containing many variables and constants. 

More recently, newer approaches to symbolic regression have arisen that make use of neural networks. The EQL (Equation Learner) model \cite{martius2017extrapolation,sahoo2018learning} is an example of performing symbolic regression by training a neural network that represents a symbolic expression. This method, and others based on it \cite{chen2020learning,kim2020integration}, take advantage of advances in deep learning as an alternative to genetic approaches. However, they still approach symbolic regression as an instance-based problem, training a model from scratch given every new input dataset for a regression task.

A recent study \cite{biggio2020seq2seq} presents a novel, language-based method for handling symbolic regression as a machine translation task, similar to the approach used by \cite{lample2019deep} for performing symbolic integration and solving differential equations. Given an input dataset, the algorithm treats the input as a text string and passes it through a trained sequence-to-sequence LSTM to produce an output text string that is parsed as the desired symbolic expression. Although this method overcomes the cost of per-instance training, its interpretation of the input dataset as a textual string limits its usability, as the input data must follow specific constraints, such as fitting a 1-dimensional mesh of fixed size. Consequently, this method can only be used in one-dimensional space. However, in most problems, more than one variable is involved and we need to find a multivariate function. In this work, we propose
a method that removes such limitations on the structure of input data. This can be applied easily to symbolic regression problems in high-dimensional spaces and when many variables are involved.

Another active area of research is to use deep reinforcement learning methods to tackle this problem \cite{kim2020integration, petersen2021deep}.
The method presented by Petersen et. al.  \cite{petersen2021deep} uses a hybrid approach between traditional genetic algorithms and deep learning methods. 
Here, the authors use deep RNNs to generate samples of candidate skeletons. As an example, if the function was $f(x) = x^2 + 1$, the corresponding skeleton would be $C_1 x^2 + C_2$. As in \cite{kommenda2019parameter}, numerical optimization is then used to optimize for the constants of each candidate skeleton. A reinforcement learning algorithm is applied to train the RNN to generate better skeletons at every iteration. However, this method still relies on the iterative nature of traditional genetic algorithms as well as numerical optimization. This results in a computationally intensive process in order to generate a prediction for each equation.

\section{Method}
\label{Method}

Our model for symbolic regression, SymbolicGPT, consists of three main stages: obtaining an order-invariant embedding of the input dataset using a T-net \cite{qi2017pointnet}, obtaining a skeleton equation using a GPT language model \cite{gpt2}, and optimizing constant values to fill in the equation skeleton. In addition to discussing each of these steps, we also present the method for generating our equation datasets.

\subsection{Equation Generation}
\label{eqn_gen}

To train our language model, we need a large dataset of solved instances of symbolic regression. This dataset is a collection of input-label pairs where each input is in the form of a numerical dataset, itself a set of input and output pairs $\left\{(\bm{x}, y)\right\}$, and the corresponding label is a string encoding the symbolic expression governing the relationship between variables in the numerical dataset.

In order to ensure that the language model is able to generalize to unseen equations, having good training data is key. 
It is necessary to train the model over a wide, diverse set of training equations in order to prevent the language model from overfitting. 

There are a number of different ways to randomly sample symbolic mathematical expressions. One approach, as used in \cite{brence2021probabilistic}, is to consider symbolic expressions as constructed by rules in a context-free grammar, and randomly sampling from rules until reaching a string containing only terminal values. 
Another approach, taken in  \cite{lample2019deep}, uses parse-tree representations of symbolic formulas, presenting a method that samples uniformly from all trees of $n$ nodes and then filling in nodes with valid operators or variable values. 

For our training dataset, we use an approach similar to the latter, where we start with a blank parse tree and then ``decorate'' the nodes with choices of operators and variables. In contrast with \cite{lample2019deep}, we do not constrain our parse trees by the number of nodes, but by the number of levels. This enables more control over the maximum level of complexity in the equations used in our training set, as the number of levels in the parse tree corresponds to the number of potential function nestings, a measure of how complex an equation can be.

We begin by fixing $k$, the maximum number of levels in the parse tree for the equations we wish to encounter in our training set. We also begin with a pre-specified number of variables, $d$, and a pre-selected set of operators, $P = \{u_1, \ldots, u_m\}$, that are allowed to appear in any training equation. Then, for each data-equation pair in our training set, we generate a perfectly balanced binary tree of depth $k$, having $2^{k-1}-1$ internal nodes and $2^{k-1}$ leaf nodes. These nodes originally start off empty to form the template of a symbolic expression.

The template is filled in by randomly selecting valid choices to occupy each node in the parse tree. For leaf nodes, each node is randomly assigned with a variable from the set $\{x_1, \ldots, x_d\}$. For interior nodes, operators from the set $P$ are randomly chosen. Once filled in, the parse tree can naturally be interpreted as a symbolic expression. For nodes filled in by binary operators, both of their child nodes are used as input; in the case of unary operators, only the left child is used as input, and the right child is ignored. Importantly, the unary operator ``id($\cdot$)'', which returns its input argument unchanged, is included in $P$, which effectively allows for equations with shallower or unbalanced parse trees to still be represented using this template.  

Additionally, to ensure that the equations generated are not all too complex, we introduce ``terminal'' nodes in which children of the terminal nodes are discarded. This ensures that we obtain a diverse set of equations within the training set. 


As a final step for the equation generation procedure, constants are incorporated into the equation by inserting them at nodes in the parse tree. Given a specified value $r \in [0, 1]$ and constant bounds $c_{min}$ and $c_{max}$, for each node in the tree, a random real-valued constant is selected between $c_{min}$ and $c_{max}$ and, with probability $r$, is inserted as a multiplicative factor the subtree rooted at that node. Similarly, a second random constant is selected between $c_{min}$ and $c_{max}$ and, with probability $r$, is inserted as an additive bias to the subtree rooted at that node. By varying the constant ratio $r$, the equations can be customized to include many constants, few constants, or none at all.

Once an equation is generated, an input dataset for symbolic regression can be produced by evaluating the symbolic expression at $n$ different vectors $\bm{x}$ randomly sampled from some region of interest in $\mathbb{R}^d$. The label value for the symbolic regression instance would be the symbolic expression. This process can be repeated many times to construct the training set by which our SymbolicGPT model will learn how to perform symbolic regression.

\subsection{Order-Invariant Embedding}


Once the training set of input data and output equations is generated, it is used to train our model for translating numerical datasets into equation strings.

The first step is to convert the input dataset $D = \{(\bm{x}_i, y_i)\}_{i=1}^n \subset \mathbb{R}^{d+1}$ into a single vector embedding $\bm{w}_D \in \mathbb{R}^e$. 
For the conversion to be useful, it must have two properties. First, it should not strictly depend on the number of points in the input dataset, $n$. In practice, the datasets provided as input to a symbolic regression solver may have varying sizes, and we do not want our method to be restricted to cases with a fixed number of input points. 

Second, the conversion method should not be sensitive to the order in which the points of the dataset are given. The input to a symbolic regression instance is a collection of datapoints, rather than a sequence, and the optimal symbolic expression to fit the dataset should not depend on the order in which the points are listed. Thus, the vector embedding of the dataset should be similarly order-invariant.

Our approach for converting the dataset $D$ into a vector embedding is to use a T-net, a kind of deep network that makes use of a global max-pooling layer to provide order-invariance over its arbitrarily-sized input \cite{qi2017pointnet}. The T-net takes as input the dataset $D$, consisting of $n$ datapoints over $d$ variables, represented in matrix format as $X \in \mathbb{R}^{n\times(d+1)}$, where $n$ can be any number and $d$, the number of allowable variables, is fixed in advance.  Any symbolic regression instance with fewer than $d$ variables can be padded with 0 values, bringing the total number of variables up to $d$.

The matrix $X$ is first normalized using a learnable normalization layer in order to regulate extreme values from the input. The normalized input points are then passed through three stages of MLP networks. Within each stage, each of the $n$ rows of $X$ are passed individually, albeit in parallel, through a single fully connected layer, where weights are shared between the networks for all points for that stage. The first stage results in $n$ points encoded in $e$-dimensional space; the second stage takes them into $2e$ dimensions, and the output after the third stage are $n$ points having $4e$ dimensions each.

\begin{figure}
  \centering
  \includegraphics[width=\textwidth]{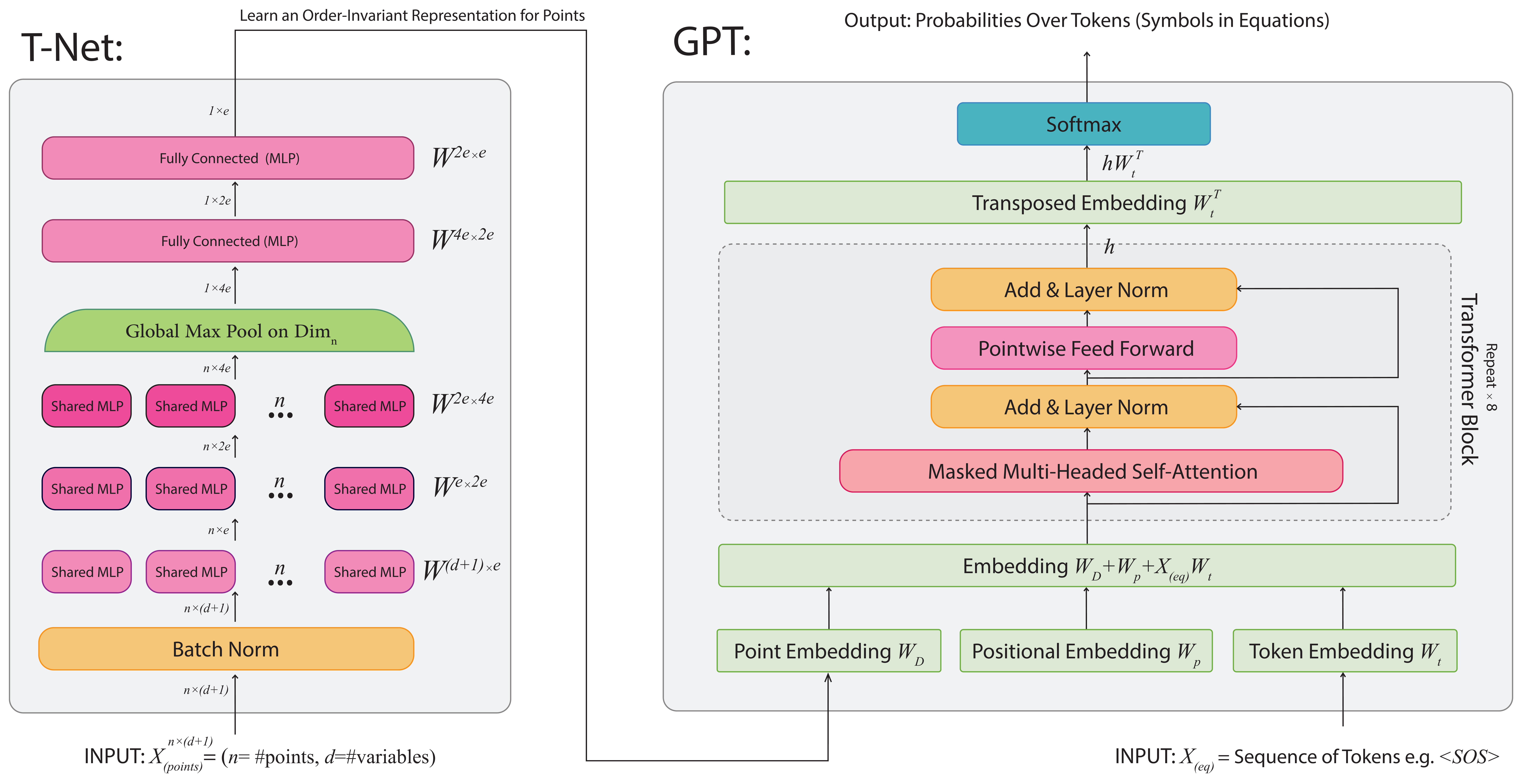} 
  \caption{The architecture of SymbolicGPT. The left box illustrates the structure of our order-invariant T-net for obtaining a vector representation of the input dataset, and the right box shows the structure of the GPT language model for producing symbolic equation skeletons.}
  \label{architecture}
\end{figure}

The next layer in the T-net is a global max pool, which reduces the $n\times 4e$ output of the previous step down to a $1\times 4e$-dimensional vector. The max-pooling eliminates the dependence on both $n$ and the order of the input points, achieving both goals needed for our vector embedding. Finally, the output of the global max-pool is passed through two more fully connected layers, resulting in a single output vector $\bm{w}_D$, an $e$-dimensional embedding of the input dataset. The overall structure of the T-net is shown in the left part of Figure~\ref{architecture}.


\subsection{Generative Model Architecture}

The main component of SymbolicGPT is the deep network for producing symbolic equations, as implemented using a GPT language model \cite{gpt1, gpt2, gpt3}. This framework takes in two pieces of input: the order-invariant embedding of the point cloud $\bm{w}_D$ as produced by the T-net, representing the input dataset, and a sequence of tokens, $X_{(eq)}$, used to initialize the output formula string. In the typical regression case where no information is provided about the output symbolic expression in advance, this token sequence would be the singleton Start-of-Sequence token $ \left<SOS\right>$, although in general it can be any desired prefix of the output equation. The input token sequence is tokenized at a character level and encoded as the matrix $W_t$ using a trainable embedding as part of the GPT model.

The first step in the GPT model is to combine the two inputs $\bm{w}_D$ and $W_t$ together, along with the positional embedding matrix $W_p$. Based on empirical support, we chose to obtain the combined embedding by taking the sum $W_p + W_D + X_{eq}W_t$, where $W_D$ is the dataset representation vector $\bm{w}_D$ expanded to fit a matrix matching the dimensions of the other embeddings.


The combined vector is then passed through $l=8$ successive transformer blocks, using the standard format of GPT models \cite{gpt2}. 
Each transformer block is a sequential combination of a masked multi-headed self-attention layer and a pointwise feed-forward network, with all blocks feeding into a central residual pathway, similar to ResNets \cite{he2016deep}.



After $l$ layers of the transformer block, the resulting output vector $\bm{h}$ is passed through a final decoder in the form of a linear projection into a vanilla softmax classifier. The projection uses the transposed token embedding matrix $W_t^{\top}$ to map the hidden state vector back into the space of tokens for symbolic expressions. The result of the softmax is a probability vector over tokens in the symbolic equation, which can be sampled to produce the best equation to describe the input dataset. 
We use top-$k$ sampling with $k = 40$ for our experiments. 

Although the symbolic equation used to generate the data can contain constant values, we do not train the GPT model to recover these values exactly. Instead, constant values in the equation are masked by $\left< C \right>$ tokens during the training phase, and the output of the GPT model is a ``skeleton equation'' which leaves these constant placeholders in the output string. This is because it is unnecessary to burden the language model with the additional task of learning precise constant values, as this can be easily handled as a separate step.

\subsection{Learning Constants}

Once the GPT model predicts a skeleton equation, we learn values of constants to decorate the skeleton as a post-processing step. This division of tasks is a common approach for string-based regression methods \cite{kommenda2019parameter, biggio2020seq2seq}.

To learning the values of constants in the symbolic equation, we employ BFGS optimization, similar to \cite{biggio2020seq2seq}, using an implementation from SciPy \cite{2020SciPy-NMeth}. The learned constant values then replace the $\left< C \right>$ placeholder tokens in the skeleton equation, resulting in the final symbolic expression to represent the given symbolic regression task.

\subsection{Evaluation Metric}

Normally, regression tasks use the mean square error as a metric for measuring the predictive accuracy of an equation. For data following equations with large values, however, this can be problematic, as the residuals can grow very large even when the predicted equation is very close to the true underlying one. To resolve this issue normalize by $\norm*{\bm{y} + \epsilon}_2$ where $\epsilon$ is used to avoid division by zero and $\norm{\cdot}_2$ is the Euclidean norm. Then the normalized mean square error, $MSE_N$, is given by 
\[
    MSE_N(\bm{y}, \bm{\hat{y}}) = \frac{1}{n}\sum_{i=1}^{n}\frac{(y_i - \hat{y}_i)^2}{\norm*{\bm{y} + \epsilon}_2} 
\]

\subsection{Strengths and Advantages}

Our method exhibits the following strengths and advantages.

\subsubsection{One-Time Training}

In contrast with most approaches for symbolic regression, our method does not start training from scratch given every new problem instance. All of the model training is performed as a one-time procedure that takes place before the GPT transformer is ever used. 
Thus, SymbolicGPT enjoys all of the benefits of allowing a pretrained model, similar to popular frameworks like BERT \cite{devlin2018bert}, which can make use of massive neural networks because the model can be trained offline in advance.

After the model is trained, every instance of symbolic regression can be solved rapidly as a problem in inference. The running time is dependent only on the initial step of reading in the input dataset, obtaining the T-Net embedding, and the final step of optimizing constant values. 

\subsubsection{GPT Technology}

Our approach to symbolic regression is based on a probabilistic language model as implemented by GPT. As state-of-the-art language models continue to evolve, our method is expected to organically improve accordingly, with no extra effort in design or implementation, by simply replacing the GPT architecture with any newer and more powerful alternative. 

\subsubsection{Scalability}

Our approach addresses two of the main problems with traditional methods.
First, our model is able to scale to multiple variables. 
Iterative methods that choose the best candidate equation at each iteration struggle as the dimension $d$ of the inputs increase since the search space of functions grow exponentially with respect to $d$.
By passing in the data points directly as inputs, the model is able to infer the dimension and produce equations accordingly. 
Second, our model is able to scale in terms of the speed in which we generate predictions. 
Existing methods that train from scratch for each regression instance are slow compared to our model. 
These methods
incrementally update their model based on computing many candidate equations.
For string-based methods, the constants within these candidate equations would need to be optimized as well. This results in a bottleneck in terms of the number of constant optimizations that need to be performed to perform inference. SymbolicGPT only performs this constant optimization once which results in significantly faster inference times.
We show empirically that SymbolicGPT produces superior results using significantly less computation time in Section \ref{results}.


\section{Experiments and Results}
\label{results}

To test our model, we implemented SymbolicGPT and trained it in a number of different settings, which we detail below. In all cases, we trained SymbolicGPT over 4 epochs using a batch size of 64. The embedding size for the T-net vector representation is $e=512$, and the maximum equation output length was capped at 200 tokens.  



Training and inference for the SymbolicGPT model were performed 
using an Intel(R) Core(TM) i9-9900K CPU @ 3.60GHz with a single NVIDIA GeForce RTX 2080 11 GB GPU and 32.0 GB Ram. It is noteworthy that our performance scores were achieved using only a single GPU, and scaling up is expected to improve training and inference times even further.


        


        



Our experimental framework consists of a large-scale comparison test where we test our model on 1000 different, randomly generated instances of symbolic regression and evaluate performance based on $MSE_N$. We repeat this test on four different settings, based on the choice of the dimension $d$: datasets with one input variable, two variables, three variables, and a random selection between one and five variables. This last setting will be referred to as the ``general'' experiment.

In each experimental setting, SymbolicGPT was trained using 10,000 randomly generated symbolic regression instances belonging to the associated dimensional configuration, each consisting of an input dataset and an equation label. A further 1000 dataset-equation pairs were generated and used for validation, and 1000 new dataset-equation pairs were generated for a test set 
he training and validation datasets used values of $\bm{x} \in [-3.0, 3.0]^{d}$ 
, and test datasets took values of $\bm{x} \in ([-5, -3] \cup [3, 5])^d$. The one, two, and three variables datasets contained 30, 200, and 500 points, respectively. The number of points in the general dataset with $d \in \{1, 2, ..., 5\}$ was a randomly selected integer between 10 and 200.

The parse tree templates, as described in Section~\ref{eqn_gen}, contained a maximum depth of $k=4$ levels and allowable operators coming from the set 
\[
P = \{\text{id}(\cdot), \text{add}(\cdot, \cdot), \text{mul}(\cdot, \cdot), \sin(\cdot), \text{pow}(\cdot, \cdot), \cos(\cdot), \text{sqrt}(\cdot), \text{exp}(\cdot), \text{div}(\cdot, \cdot), \text{sub}(\cdot, \cdot), \log(\cdot)\}
\]
Constant values selected from the interval $[-2.1, 2.1]$ 
were randomly inserted using a constant ratio $r = 0.5$. 

We compared our methods  with three existing models for nonlinear regression:

\begin{enumerate}
    \item Deep Symbolic Regression (\textbf{DSR}): We use the method in \cite{petersen2021deep} to represent the most recent developments in deep learning methods for symbolic regression. The DSR algorithm can be very effective for simple equations, but includes a constant optimization step that is extremely expensive for larger configurations, computationally. In order to complete experiments within a reasonable running time, we limited the population size to be 1000 and trained for a maximum of 10 epochs.
    
    \item Genetic Programming (\textbf{GP}): We chose to use Python's GPLearn package to represent genetic evolution algorithms for symbolic regression. We use two models with different configurations for this experiment. We refer to \textbf{GP} as the model with a population size of 1000 and 10 generations. \textbf{GP Max} is the model with a population size of 5000 and 20 generations.
    
    \item Neural Network (\textbf{MLP}): We use a standard Multilayer Perceptron to act as a non-symbolic, nonlinear regressor to use as a baseline for comparison, as implemented in the Python package Scikit-Learn \cite{JMLR:v12:pedregosa11a}.
\end{enumerate}

\begin{figure}
    \centering
    \includegraphics[width=\textwidth]{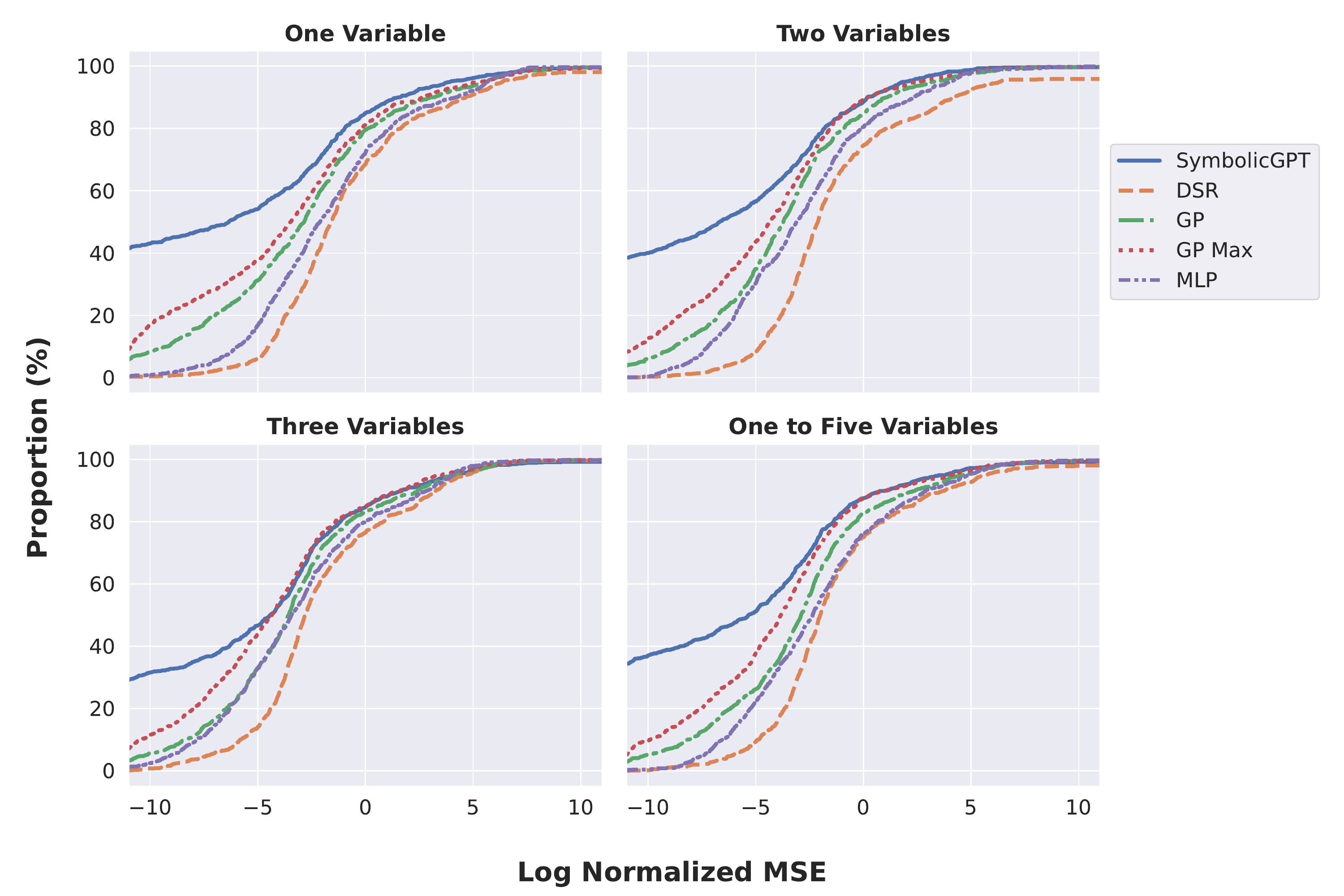}
    \caption{Cumulative $\log MSE_N$ over all methods and experiments. Each curve shows the proportion of test cases that attained an error score less than every given threshold. SymbolicGPT finds better fitting equations for more test cases than DSR and finds more highly accurate equations (with $\log MSE_N < -10$) than any other method tested. }
    \label{fig:all_combined}
\end{figure}

\begin{figure}
    \centering
    \includegraphics[width=\textwidth]{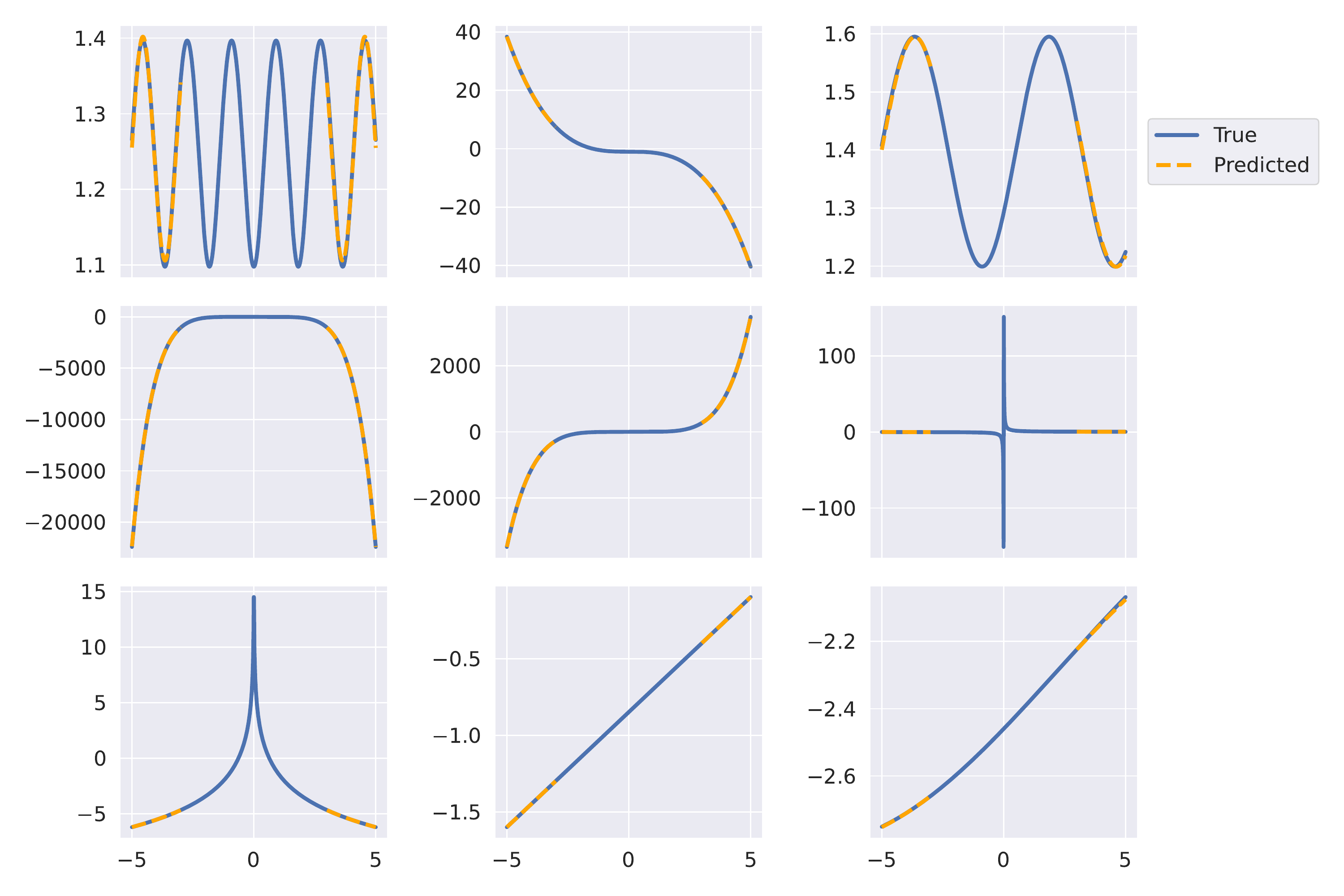}
    \caption{Graphical representations of selected equations of one input variable. The solid blue curves are the graphs of the true underlying equations; the orange dotted curves are the predicted functions as generated by SymbolicGPT.} 
    \label{fig:cherry_pick}
\end{figure}

For each method, we evaluated its performance on 1000 test instances of symbolic regression in each of the four experiment settings, using $MSE_N$ as the fitness metric. We summarized the results in the cumulative distribution plots of Figure~\ref{fig:all_combined}, showing the proportion of the test cases that attained error less than any given threshold value. Methods corresponding to curves positioned higher in the plot achieved higher accuracy on more test equations, and hence are better regressors. However, the most important region of the plot is the far left side, as the number of test cases that achieved the lowest possible error is an indication of how often the method would find a highly accurate fitting equation. 
Some visualized examples of predictions generated by SymbolicGPT are presented in Figure~\ref{fig:cherry_pick}. 

To give an indication of the speedy performance of SymbolicGPT, we measured and compared the average running time to solve an instance of symbolic regression in the general experiment setting. The mean running times, along with standard deviations, are shown in Table~\ref{tab:running_times}.
In order to make a fair comparison between running times, all experiments were performed using the same computer specifications.
We excluded the MLP regressor from this experiment in order to compare between strictly symbolic regression methods. The results show that SymbolicGPT requires significantly less time to solve an instance of symbolic regression compared with other methods, often by an order of magnitude or more, due to most of the computation being shifted to the offline step of setting up the pre-trained model. 

\begin{figure}
    \centering
    \includegraphics[width=0.9\textwidth]{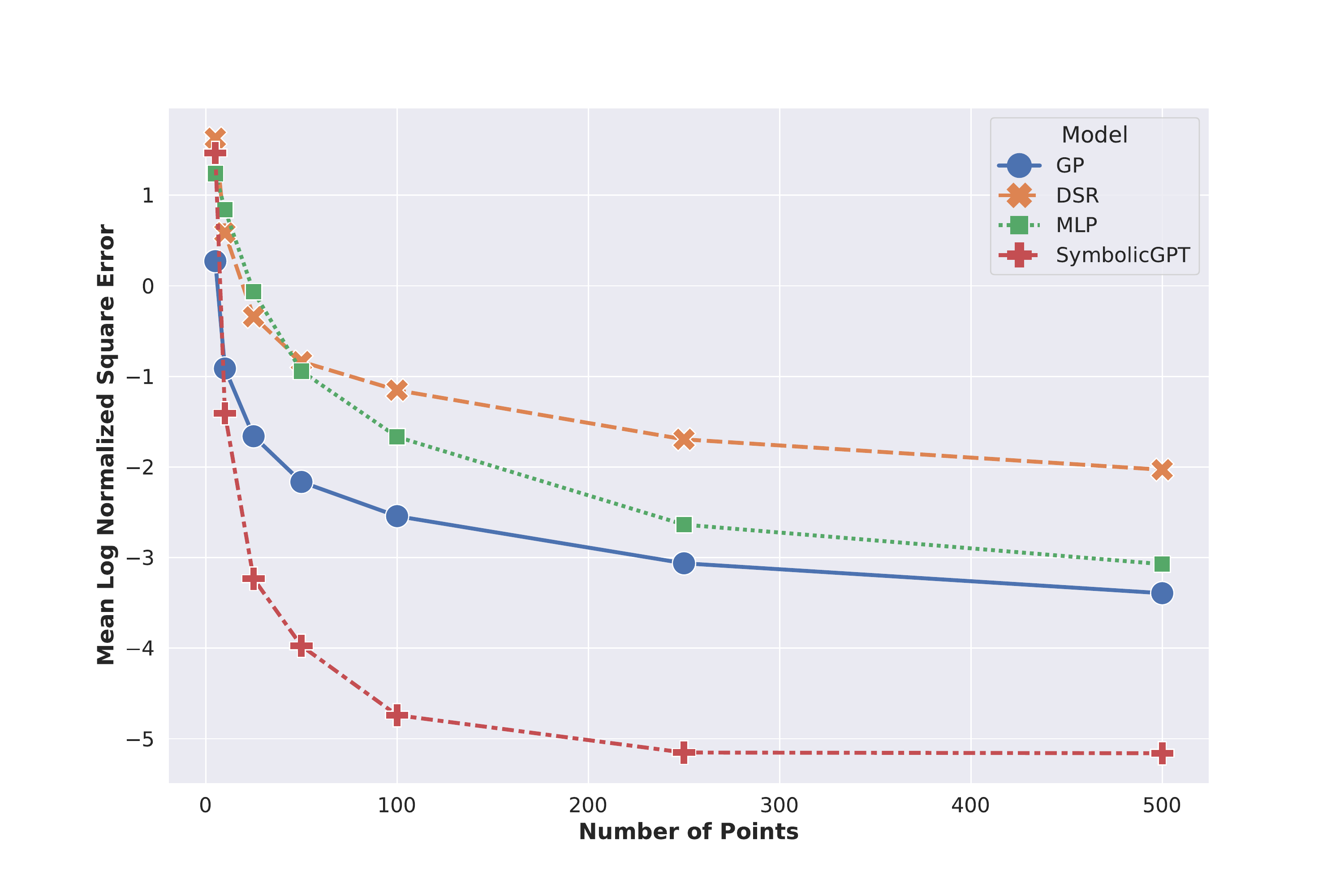}
    \caption{The effect of the number of points on the performance of the model. }
    \label{fig:n_points}
\end{figure}

We also ran each of the four regression algorithms on symbolic regression instances with the varying number of input datapoints, in order to gauge the data efficiency of each of the methods. The results of this experiment are shown in Figure~\ref{fig:n_points}. As the plot indicates, all methods improve performance (by reducing regression error) as more training points are provided; however, SymbolicGPT consistently achieves lower error scores than all other methods, regardless of how many data points are available in the symbolic regression instances. In particular, SymbolicGPT, when given datasets of just 50 points,  achieves lower regression error than the other algorithms do on instances with up to 500 points. This is in spite of the fact that the SymbolicGPT model was trained only on datasets of 500 points each: the robustness to differently sized input dataset instances is possibly a consequence of the order-invariant T-embeddings used in the model.

\begin{table}
\centering
\begin{tabular}{@{}lllll@{}}
\toprule
Experiment   & GP       & GP Max         & DSR               & SymbolicGPT          \\ \midrule
General      & $48.0 \pm 26.7$ & $84.8 \pm 25.8$ & $78.8 \pm 42.8$ & $\bm{5.0} \pm 12.0$     \\
One variable & $44.6 \pm 33.0$ & $82.1 \pm 32.1$ & $15.1 \pm 2.1$  & $\bm{1.1} \pm 0.9$ \\
Two variables   & $47.3 \pm 29.4$ & $100.8 \pm 31.9$ & $76.7 \pm 39.8$ & $\bm{3.5} \pm 9.0$   \\
Three variables & $60.0 \pm 32.9$ & $109.5 \pm 32.4$ & $73.3 \pm 56.4$ & $\bm{10.3} \pm 26.2$ \\ \bottomrule
\end{tabular}

\caption{Average running times (in seconds) for an instance of symbolic regression during each of the four experiments.}
\label{tab:running_times}
\end{table}

\section{Conclusions}
\label{conclusions}
In this work, we have presented a method that pushes the boundaries of language models and approaches the problem of symbolic regression from a new and powerful direction. We have employed language models in a novel way and with a novel approach, combining them with symbolic mathematics and order-invariant representations of point clouds. Our approach eliminates the per-instance computation expense of most regression methods, and resolves the input restrictions imposed by other language-based regression models. Moreover, our method is fast, scalable, and performs competently on several kinds of symbolic regression problems when compared with existing approaches.




\bibliographystyle{abbrvnat}
\bibliography{main}

\end{document}